\documentclass{article}
\usepackage{graphicx}
\usepackage{amssymb}
\usepackage{amsmath}
\usepackage{algorithm}
\usepackage{algorithmic}
\usepackage{amsfonts}
\usepackage{epstopdf}
\usepackage{color}
\usepackage[square,numbers,sort]{natbib}

\usepackage[utf8]{inputenc} 
\usepackage[T1]{fontenc}    
\usepackage{hyperref}       
\usepackage{url}            
\usepackage{booktabs}       
\usepackage{amsfonts}       
\usepackage{nicefrac}       
\usepackage{microtype}      

\usepackage[final]{nips_2018}

\title{The Barbados 2018 List of \\Open Issues in Continual Learning} 
\author{
Tom Schaul \hspace{3.5em} Hado van Hasselt \hspace{2.5em} Joseph Modayil 
\\ {\bf
 Martha White \hspace{2.5em} Adam White \hspace{4em} Pierre-Luc Bacon \hspace{3em} Jean Harb 
} \\ {\bf
Shibl Mourad \hspace{3.5em} Marc Bellemare \hspace{3.5em} Doina Precup 
\vspace{-1.em}
}}

\begin{document}
\maketitle

\section*{Motivation}
We want to make progress toward artificial general intelligence, namely general-purpose agents that autonomously learn how to competently act in complex environments.
The purpose of this report is to sketch a research outline, share some of the most important open issues we are facing, and stimulate further discussion in the community.
The content is based on some of our discussions during a week-long workshop held in Barbados in February 2018. 

\section*{Setting}
We adopt the reinforcement learning (RL) formulation, where an agent interacts sequentially with an environment, and the agent is provided a reward signal that  unambiguously defines success.\footnote{According to the \emph{reward hypothesis}, this is sufficient to represent all goals or purposes \cite{sutton2018reinforcement}.}
We want to explicitly consider some of the most challenging dimensions for a developing intelligence.
\begin{description}
\item[Vast worlds:] The environment can be much larger than the agent's capacity to move around and absorb information, so only a tiny fraction of states are ever visited.
\item[Firehose:] The stream of observations can be much richer than what the agent can store.
\item[Poor reward signal:] The reward signal may provide little information, with rare signs of progress.\footnote{Or worse,  deceptive rewards for which myopic local maximization is at odds with good long-term behaviour.}
\item[Non-stationarity:] The environment's dynamics and reward function may be changing.\footnote{Vast stationary worlds can  appear non-stationary to an agent with limited resources.
}
\item[Single continuing life:] The agent is never reset.  It has a single uninterrupted stream of experience. \item[Irreversibility:] The agent's choices can irreversibly prune its possible futures. 
\end{description}
We hypothesize that environments with these properties will require agents that \emph{flexibly learn to predict and control their data stream}, or in other words, agents that form exploitable knowledge about their environment in a continual, constructive, and accumulating manner. 

\section*{What is Continual Learning?}
While the RL problem formulation with its reward-maximizing objective remains unchanged, we consider a class of problems that require generalization and exploration in a complex setting to maximize rewards. This class of problems suggests a class of solution methods that embody the particular inductive bias that accumulated knowledge (representations, experience, skills, predictions) can be useful for maximizing reward in the present. It introduces a trade-off between learning for the future (new knowledge) and acting for the present (maximizing reward only).
Perhaps this trade-off needs to be chosen in advance, for instance by encouraging the learning agent to keep looking for something better, and not to converge to a stationary behaviour, in case the world is non-stationary or better rewards can be found.  It may not be possible to learn this trade-off from the provided data alone---perhaps this needs to be an inductive bias of the learning system.
We call this class of problems \emph{continual learning}~\citep{ring1994continual}, and note that it is closely related to learning in humans and animals, and roughly synonymous with `life-long learning'~\citep{ruvolo2013ella}.

\section*{Open Issues}
We now list what we consider the critical issues for continual learning. We emphasize those that are particularly challenging for continual learning, as opposed to those that affect all of RL. Some of these may be relevant outside of a continual learning (or even RL) setting as well. 

\subsection*{1. Evaluating continual learning} 
The first top-level challenge is how to guide the research on continual learning. We believe that evaluation is instrumental to rapid progress, clear communication, as well as reproducibility and comparability of solution methods.  

\subsection*{1.1. Domains}
We need domains that exhibit the kinds of challenges that make continual learning beneficial (ideally in a decoupled and scalable way) and support measurable progress and meaningful comparisons between algorithms. Ideally we would have sets of domains that include simple understandable illustrative problems that highlight core issues, as well as vast complex worlds in which some strong form of continual learning is necessary for meaningful progress.

\subsection*{1.2. Metrics}
The appropriate metric of success is not always immediately clear.  Conventional metrics include overall reward during a lifetime and average reward per step, but these might not be the most informative for the researcher, or for the learning agent.
To enrich the signal for researchers, it could be valuable to have a set of established auxiliary metrics of continual learning that go beyond the increase in reward. Those could be in the form of \emph{probe questions} on which the accuracy of agent's answers is measured from time to time. In addition, we may need to embrace qualitative measures, such as behaviour analysis to measure progress. 

\subsection*{2. Designing continual learning agents}
The second top-level challenge is how to build agents that learn to predict and control the data stream, by effectively \emph{integrating} the solutions to the challenges below. We expect that it will be unlikely for each challenge to be solved entirely in isolation.

\subsection*{2.1. Choosing what to learn about (discovery)}
The agent will formulate a set (or distribution) of \emph{questions} about its data stream. We posit that these questions should be \emph{predictive}, allowing the agent to autonomously \emph{verify} the quality of its answers~\citep{degris2012scaling}. 
The choice of predictive questions is rooted in the observation that predictions are versatile and powerful---if we would be able to predict any desirable quantity or event as well as possible, this would seem sufficient to encode most relevant knowledge for optimal behaviour.
The challenge of discovery is determining what those questions should be, and has the following sub-problems.

\paragraph{2.1.1. Representing questions}
We consider \emph{general value functions} (GVF) to be a promising formalism for representing the questions~\citep{sutton2011horde}. GVFs are agent-state-dependent (and, potentially, action-dependent) predictions conditioned on a policy, a signal that is accumulated (e.g., reward), and a temporal horizon (e.g., discount). These GVFs generalize the value function in conventional RL, in the sense that the agent can maintain many GVFs in parallel, each attending to different signals and horizons. In addition to predicting GVFs, the agent could form \emph{control} questions that ask for a policy which optimizes a given signal.

\paragraph{2.1.2. Generating questions} 
An agent's central question should include what is the expected continual (or average) reward.  Auxiliary questions can be constructed bottom-up by identifying other signals, and predicting (or controlling) accumulations of those signals~\citep{littman2002predictive,Schaul2013forecasts}.
Auxiliary questions can also be constructed top-down, by asking predictive questions when the agent has a use for the answers.  Classical model-based reinforcement learning provides an example: a one-step transition model and an immediate reward model are both examples of predictions, and are special cases of GVFs.  
There are many possibilities between a full bottom-up approach and a full top-down approach. 

\paragraph{2.1.3. Assessing questions}
The fact that there are many different ways to generate questions suggests that criteria are necessary to judge the utility of questions, possibly before even knowing how to answer them.
The agent's reward itself should not be the only criterion.
For example, a good new question could be one that helps the agent to better answer its previous set of questions.

\subsection*{2.2. Gathering the experience needed to learn (exploration)}
The agent has partial control over its incoming data stream. Its challenge is to both be \emph{effective} and make the \emph{right trade-offs}. For example, an agent needs to trade-off novelty and chance of survival, and maximize the quality of its experience for a given risk profile. 
We hypothesize that it is not enough to have these decisions be guided purely by reward (even if that is the ultimate role of exploration), and propose three dimensions of the exploration problem: drives, skills and trade-offs.
Some exploration principles for single-task settings, such as optimism in the face of uncertainty, may not be suitable for a continual learning setting, if they encourage exhaustive exploration when the world is too big to cover fully.

\paragraph{2.2.1. Drives}
The agent could pursue drives or `intrinsic motivations', which are separate from (extrinsic) reward, but are useful to gain experience that helps obtain reward later.
An intrinsic motivation could be a scalar signal, an intrinsic reward, such that an agent maximizing it 
would keep exploring to learn new things.
Drives can be generic inductive biases that include short-horizon objectives such as seeking novel actions, observations or states, and longer-horizon objectives such as discovering states that are difficult to reach, states where the consequences of the agent's actions are large and known (controllability),  
or states from which the agent can quickly/cheaply get to many other states (empowerment~\citep{mohamed2015variational}).
Other drives can be more agent-focused, 
by depending on the current set of questions and the agent's current knowledge about them. Examples include learning progress (`learning feels good'), compressibility, or uncertainty reduction in the answers~\citep{schmidhuber1991possibility}.

\paragraph{2.2.2. Building exploration skills}
The agent should gradually develop the skills to more effectively learn answers to \emph{new} questions. This setting could benefit from additional mechanisms and inductive biases (besides intrinsic motivations), e.g., commitment/persistence, or consistent/repeated behaviour. 

\paragraph{2.2.3. Survival and risk-awareness}
A separate issue from defining drives and skills is knowing how to trade them off against each other and against gathering extrinsic reward. One difficulty particular to continual learning is the sensitivity of a single-life agent to death, and thus to taking risks. To properly handle risk, it may be necessary for an agent to estimate more than a mean outcome~\citep{bellemare2017distributional}.

\subsection*{2.3. Learning in preparation for future questions (knowledge)}
The way to act competently in a non-stationary world is to generalize effectively from past experience to the present. This challenge has multiple related but distinct dimensions.

\paragraph{2.3.1. Representing many answers}
The issue of how to best represent the answers to many questions points to a spectrum, with non-parametric (each question is answered separately) and non-interfering architectures on one end~\citep{sutton2011horde}, and parametric architectures that generalize across questions on the other end~\citep{uvfa2015}, with more modular approaches in-between.

\paragraph*{2.3.2. Learning many answers}
We can learn \emph{off-policy} to answer questions conditioned on policies other than the current behaviour. How best to learn efficiently and off-policy remains an open issue, although a panoply of off-policy learning methods exists~\citep{munos2016safe}. 

\paragraph{2.3.3. Learning from few examples} 
In continual learning we cannot assume that all states are visited infinitely often.   With non-stationarity it is more appropriate to assume that only a tiny fraction of states is ever encountered. The consequence for learning is that data is \emph{precious}, and should be used in the most effective way possible. 

\paragraph{2.3.4. Appropriate generalization}
For efficient learning, we need to generalize across states, across questions, and across time, from the past to the future. Unlike the i.i.d.~setting for generalization in supervised learning, in continual learning there is a fundamental issue of non-stationarity: it may be impossible to tell with certainty whether the world has changed forever (e.g., because of an irreversible decision), or whether a change is periodic.
Even in the latter case however, a capacity-limited agent has to balance how much to learn from new experience and how much old knowledge to forget or overwrite (stability-plasticity dilemma~\citep{abraham2005memory}).

\paragraph{2.3.5. Balancing learning effort and inference effort} 
In a vast world, agents are always capacity-limited, both in terms of memory and compute. Furthermore, the available compute is used both for learning and for inferring the best next action, leading to a trade-off between these two. A flexible use of memory capacity allows the agent to cache some of the decision-making computation (e.g., using action-value functions instead of re-planning every step).

\paragraph{2.3.6. Trading off multiple objectives} 
In continual learning, more than in other settings, it may be critical not to maximize one objective at the expense of all others. 
It is unclear whether the most effective approach involves collapsing multiple competing objectives to a single one via a fixed weighting---alternatively an explicit multi-objective formulation~\citep{roijers2017multi} may be appropriate.

\subsection*{2.4. How to use knowledge}
It can be non-trivial to use learned knowledge about non-reward quantities to improve the agent's effectiveness at obtaining reward.

\paragraph{2.4.1. Using knowledge to improve learning}
Learning representations (e.g., features) that are shared between many questions and answers is useful for generalization, as they capture useful qualities of the world, as well as for exploiting reward~\citep{jaderberg2016reinforcement}. Alternatively, the answers to some questions can be used as features that help learn other answers efficiently. 

\paragraph{2.4.2. Using knowledge to improve behaviour}
The knowledge of how to control or optimize many signals in the environment can be used for more directed behaviour than dithering on low-level motor controls, which in turn can help obtain interesting experience or reward.

\paragraph{2.4.3. Using knowledge to infer more knowledge}
If an agent could model the full world, it would not have to spend any more time exploring the world: it could plan an optimal solution.  Even when this cannot be done exactly, because the world is too large and not all relevant parts can be observed, this principle can still apply.  The best ways to build models and plan with them remains an open issue.  We expect the common approach of building one-step models that  predict the raw observations at the next step to be insufficient: such models allocate resources to irrelevant signals and iterating such low-latency models over long horizons can easily compound small modelling errors into large errors. Classical planning algorithms assume the models are flawless.  New planning algorithms may be necessary that yield meaningful results when used with inaccurate or partial, or even abstract, learned models~\citep{silver2016predictron,mctsnets}.

\section*{Conclusion}
This document provides our view of open issues where progress is possible.  We hope this list provides a fruitful starting point for researchers wishing to study continual learning by building upon existing ideas and algorithms from reinforcement learning.

\subsection*{Acknowledgements}
We thank Richard Sutton and David Silver for inspiring and shaping the discussions, as well as Thomas Degris and Andre Barreto for their constructive comments on an earlier draft. 

\bibliographystyle{abbrvnat}
\bibliography{bib}

\end{document}